\documentclass[letterpaper, 10pt, conference]{ieeeconf}  

\IEEEoverridecommandlockouts                              

\usepackage[T1]{fontenc}
\usepackage[table]{xcolor}
\usepackage{booktabs}
\usepackage{multirow}
\usepackage{graphicx}
\usepackage{float}
\usepackage{url}
\usepackage{amsmath}
\usepackage{subcaption}

\definecolor{bestRed}{HTML}{F8CECC}   
\definecolor{worstYellow}{HTML}{FFFFBF} 
\definecolor{visualcolor}{RGB}{245,245,245}
\definecolor{depthcolor}{RGB}{235,245,255}
\definecolor{normalscolor}{RGB}{240,250,240}
\definecolor{geometriccolor}{RGB}{255,248,235}

\title{\LARGE \bf
Evaluating Depth-Augmented Visual Features for Cross-Dataset Lunar Traversability Estimation*
}

\author{Jan Baturo$^{1,2}$ and Bartosz Ptak$^{2}$
\thanks{*This work was not supported by any organization.}
\thanks{$^{1}$Baturo J. is with European Space Agency (ESA Spaceship Poland), Poznań, Poland.}%
\thanks{$^{2}$Baturo J. and Ptak B. are with Institute of Robotics and Machine Intelligence, Poznań University of Technology, Piotrowo 3A, 60-965 Poznań, Poland
        ({\tt\small bartosz.ptak@put.poznan.pl}).}%
}

\begin{document}

\maketitle
\thispagestyle{empty}
\pagestyle{empty}

\begin{abstract}
Autonomous mobility is essential for sustained lunar surface operations, as communication delays and line-of-sight occlusions limit continuous teleoperation. Although recent traversability estimation methods increasingly rely on deep learning, the domain gap between terrestrial analogue environments and real lunar terrain can limit their generalisation, and real lunar traversability labels are scarce, making fully supervised training difficult. To address these challenges, we build a weakly supervised framework trained on sparse pairwise annotations and study whether augmenting visual features with depth-derived geometric information: depth, surface normals, and combined geometric descriptors. We evaluate the models in both in-domain and cross-domain settings using two terrestrial analogue datasets and the real lunar Chang'E PCAM dataset. Contrary to expectations, we find that naive geometric augmentation yields no consistent improvement over a strong visual-only baseline in either setting, with effects small relative to seed-to-seed variance. Instead, the choice of training source emerges as the dominant factor in transfer quality: models trained on domain-relevant sources transfer substantially better to real lunar imagery than the geometric fusion strategy improves any single model. Our evaluation runs onboard a resource-constrained edge device and provides a controlled baseline for designing more expressive multimodal fusion architectures in future lunar perception systems.
\end{abstract}


\section{Introduction}

Autonomous mobility is required for long-term lunar surface operations. Communication latency and frequent line-of-sight occlusions limit continuous teleoperation, so a rover must judge for itself whether the terrain ahead is safe to cross. Regardless of the rover's mode of operation -- whether it operates autonomously or is controlled remotely -- terrain traversability estimation can increase movement speed and reduce hazardous situations. Recent traversability methods increasingly build on learned visual features, and self-supervised foundation models such as DINOv2~\cite{oquab2023dinov2} produce dense descriptors that transfer across domains far better than encoders trained from scratch~\cite{schreiber2025you}. Lunar imagery, however, is strongly out-of-distribution: illumination is harsh and unfiltered, and surface albedo and regolith texture differ from terrestrial simulants and between analogue facilities. Even foundation features, therefore, retain a residual domain gap, and how well they transfer to traversability across lunar datasets is seldom measured~\cite{ptakearth}. Although the traversability investigated in this work has no physical reference, it provides semantic understanding of unstructured terrain, increasing the rover's ability to distinguish safe from hazardous regions based on visual appearance alone.

\begin{figure}[b!]
  \centering
  \includegraphics[width=\linewidth]{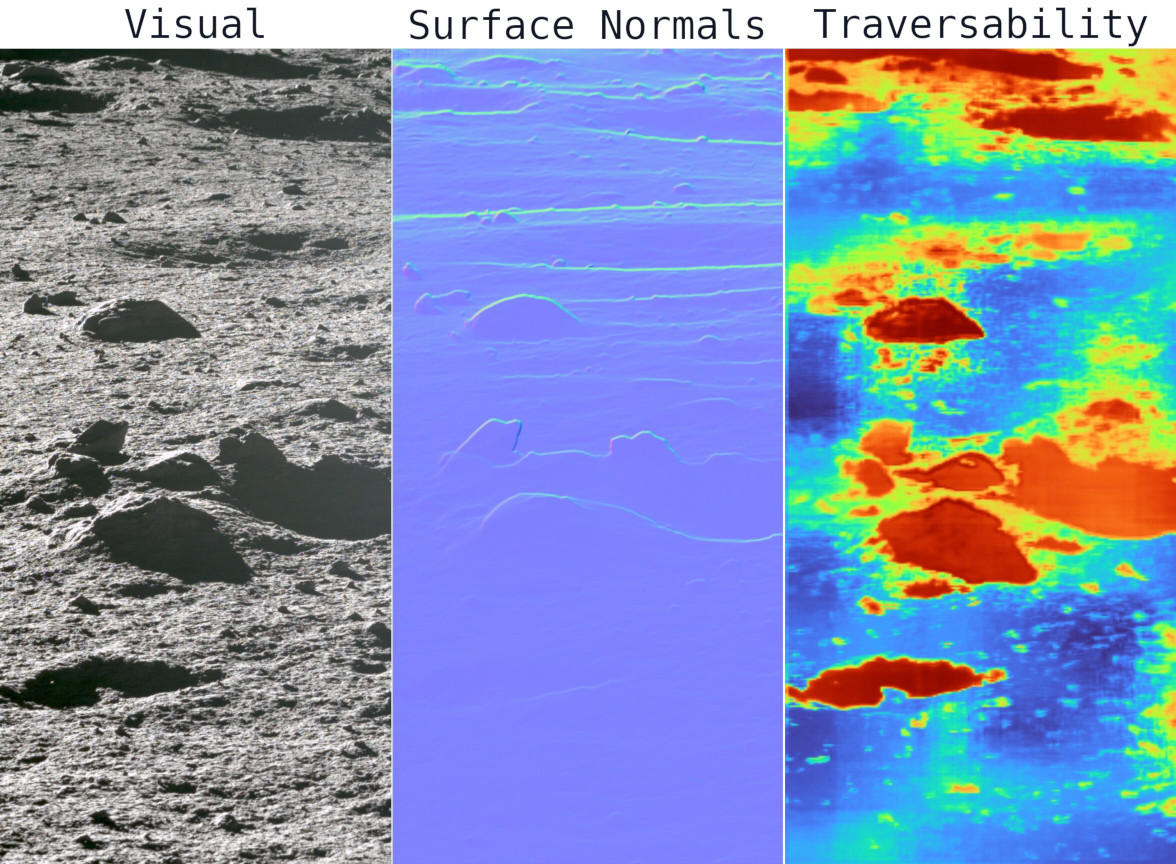}
  \caption{The same real lunar scene (Chang'E PCAM) shown as visual, surface normals, and predicted traversability. Photometric appearance shifts markedly across domains, while the geometry governing traversability is largely preserved, motivating depth-derived inputs for cross-dataset transfer.}
  \label{fig:teaser}
\end{figure}

Recent weakly supervised approaches estimate traversability by extracting dense per-pixel descriptors through FeatUp upsampling of DINO features and ranking terrain traversability using a lightweight MLP that operates on visual imagery alone. We hypothesise that augmenting these visual representations with explicit geometric information -- namely depth, surface normals, and handcrafted geometric features -- can further reduce the remaining cross-domain gap. Unlike photometric cues, which vary with illumination conditions and sensor characteristics, the physical determinants of traversability, such as slope, surface roughness, and the geometry of rocks and crater rims, remain largely invariant across domains (Figure~\ref{fig:teaser}). Building on this paradigm, we extend visual feature-based traversability estimation by integrating upsampled DINO descriptors with geometric channels. We train the resulting model within the weakly supervised W-RIZZ approach~\cite{schreiber2024w} and evaluate it using the Human Disagreement Rate (HDR).

Our specific contributions are:

\begin{itemize}
    \item an adaptation of CHUNGUS (Continual Human Understanding for Navigational Guidance in Unstructured settings)~\cite{schreiber2025you} with geometry-augmented inputs -- depth, surface normals, and handcrafted terrain descriptors;

    \item a cross-dataset evaluation protocol spanning two lunar analogue facilities (LunaPolaris~\cite{van2026lunapolaris} and our in-house OTRK Kąkolewo) and real lunar imagery from the Chang'E-3/4 panoramic cameras~\cite{gras_ce3_pcam,gras_ce4_pcam}, reporting all train--test domain pairs;

    \item feasibility of the method working on an edge device.
\end{itemize}


\section{Related works}

The literature generally categorises traversability methods into 3 main scopes: geometry-based, vision-based and fusion-based. Geometry-based approaches mostly rely on using exteroceptive sensors such as LIDAR and depth cameras. Firstly applied techniques were computationally lightweight and straightforward to implement, however, they often lacked robustness and struggled to adapt to unstructured, rough terrains. Advancements in machine learning and deep neural networks encouraged researchers to estimate traversability from point clouds by applying machine learning techniques~\cite{lidar_trav_1,lidar_trav_3,lidar_trav_4}. Nevertheless, relying exclusively on geospatial data presents significant disadvantages as it ignores semantic properties of the terrain -- such as distinguishing between solid bedrock and highly compressible sand. Additionally, the high cost and computational demands of LiDAR-based algorithms led to the development of vision-based methods.  These approaches used various architectures, e.g., AlexNet or PSPNet, and applied different training methods, such as contrastive, self-supervised, or supervised learning~\cite{rgb_trav_1,lidar_trav_2,rgb_trav_3}. As vision-based methods relied solely on visual appearances, fusion-based methods were subsequently developed to incorporate multiple sensory modalities, creating more robust traversability representations. One of the methods developed using multi-modal fusion and tested for space environments is Terrain Traversability Analysis using Geometry and Environmental Perception (TTA-GEP) for planetary rover path planning. The TTA-GEP system architecture operates on highly parallelised dual-stream processing, merging metric data with general object knowledge into one cohesive whole. The geometric stream uses stereovision to generate point clouds, which are processed into a 2.5D map that can be used to extract traditional geometric features such as slope or terrain roughness. Simultaneously, a powerful terrain classifier based on deep learning architecture processes images in order to semantically segment the environment. The goal is to accurately classify every segment of the Martian or lunar landscape. Semantic labels applied in TTA-GEP research divide the environment into categories highly specific to planetary robotics: bearing soil, exposed bedrock, fine sand, loose rocks, crater edges and ridges, hazardous depressions, and sky background. The last part is the continuous evaluation function, which mathematically fuses and weighs on-the-fly the cost derived from local terrain geometry with the discrete material label derived from semantic segmentation~\cite{TTA-GEP}.



\section{Method}


Our traversability algorithm is based on the approach presented in the article “Do You Know the Way? Human-in-the-Loop Understanding for Fast Traversability Estimation in Mobile Robotics”\cite{schreiber2025you}. The original method (called CHUNGUS) combines a self-supervised visual backbone with a lightweight prediction head to estimate terrain traversability from sparse human annotations~\cite{schreiber2024w}. It uses a pre-trained DINO model to extract general visual features from RGB images. FeatUp upsamples these embeddings, then passes them to a lightweight multilayer perceptron (MLP) network that predicts terrain traversability. In our method, we concatenate geometric features directly to the DINOv2 embedding vector as raw values, yielding a (384 + d)-dimensional input to the MLP head. We choose this naive fusion strategy to preserve the frozen backbone's feature space without retraining and to evaluate the contribution of geometric cues while minimising changes to the network architecture and their influence compared to the visual-only baseline. More sophisticated fusion mechanisms, such as cross-attention or learned gating, are left for future work.

We train the model using sparse pairwise annotations. Instead of assigning absolute values, a human annotator selects two points and specifies their relative traversability (less traversable, more traversable or equally traversable) -- see labelling process details in Section~\ref{sec:datasets}.

The original framework included a Human-In-The-Loop (HITL) component that required additional annotations when the robot encountered unfamiliar terrain during online operation. Since we evaluate geometric features' influence in offline behaviour, we removed the human-in-the-loop component to preserve a fair comparison. The model trains entirely offline, simplifying deployment on edge devices. Compared to the original approach, we also replaced FeatUp\cite{fu2024featup} with AnyUp\cite{wimmer2025anyup} because it better preserves semantic consistency, aligns spatial features more sharply during upsampling, and is more compatible with modern deployment libraries.

\subsection{Geometric information}

To assess the integration of deep visual features and geometric information, three depth-based representations were considered:

\begin{itemize}
    \item \emph{Depth} - DINO features are concatenated with a depth channel normalised to the [0,1] range.

    \item \emph{Normals} - Added 3 channels (x, y, z) encoding local surface orientation, capturing slope and discontinuities. 

    \item \emph{Geometric} - In addition to the surface normals, three supplementary terrain descriptors were incorporated: 
\begin{itemize}
    \item Slope magnitude: $\sqrt{(\partial_x d)^2 + (\partial_y d)^2}$, which quantifies the local steepness. Here, $d$ denotes the local depth value, while $\partial_x d$ and $\partial_y d$ are the partial derivatives (gradients) of the depth in the $x$ and $y$ directions, respectively.
    \item Local roughness: depth standard deviation in a ($5 \times 5$) neighbourhood (denoted as $\sigma$).
    \item Z-Score anomaly: ($z = (d - \mu)/\sigma$), used to identify local depth outliers, where $z$ is the resulting anomaly score, $d$ is the depth at the point of interest, $\mu$ is the mean depth of the surrounding local neighbourhood, and $\sigma$ is the standard deviation of the depths within that same neighbourhood.
\end{itemize}
    
\end{itemize}
\noindent
These additional geometric features are concatenated with the upsampled DINO feature embeddings for every pixel before passing them to MLP. This combined representation allows the network to leverage both semantic and geometric features.

\subsection{Implementation details}

Each depth image was processed independently. We filled invalid measurements, defined as non-positive or non-finite values, using Navier--Stokes inpainting with a radius of three pixels. We used the valid measurements to determine the 1st and 99th percentile bounds, then clipped the depth values to this range and min--max normalised them to approximately $[0,1]$. A bilateral filter was subsequently applied. Each geometry channel was independently scaled per image to $[0,1]$. We applied no dataset-level mean or standard deviation normalisation to the depth or geometry channels. We normalised only the visual input using the ImageNet mean and standard deviation.

For the in-domain experiments, we divided each dataset into training, validation, and test subsets using a 70/15/15 split. For the cross-domain experiments, we divided the source dataset into 70\% training and 30\% validation subsets, while reserving the full target-domain dataset exclusively for testing. We based early stopping and best-checkpoint selection solely on validation data from the source domain. We used no samples or evaluation results from the cross-domain test set during training, early stopping, or model selection.

To train the prediction head, we used a similar strategy to Schreiber et al.\cite{schreiber2024w}, adopting $L_{Rizz}$ loss and training on sparse annotations. We resize input images to 224×224 and process them with a pre-trained DINOv2 ViT-S/14 backbone to produce 384-dimensional feature embeddings. We used a fixed seed for reproducibility. Optimisation is performed with a learning rate of ($3 \times 10^{-3}$) and zero weight decay. We decay the learning rate by a factor of 0.5 every 75 epochs. Training runs for up to 200 epochs, with early stopping enabled and a patience of 75 epochs based on the validation metric, HDR@0.25.


\begin{table*}[ht!]
\centering
\caption{Summary of datasets used in this work.}
\label{table:dataset_comparison}
\resizebox{0.7\textwidth}{!}{%
\begin{tabular}{|c|c|c|c|c|}
\hline
\textbf{Dataset} & \textbf{Source} & \textbf{Depth} & \textbf{Images} & \textbf{Labels} \\ \hline
LunaPolaris & Analogue & Stereovision & 2404 & 3606 \\ \hline
In-House & Analogue & Indirect ToF & 2402 & 3603 \\ \hline
CE'3/CE'4 & Moon & Stereovision (estimated using neural methods) & 566 & 849 \\ \hline
\end{tabular}%
}
\end{table*}

\section{Experiments}

\subsection{Datasets}\label{sec:datasets}

\begin{figure}[t]
    \centering
    
    \includegraphics[width=\columnwidth]{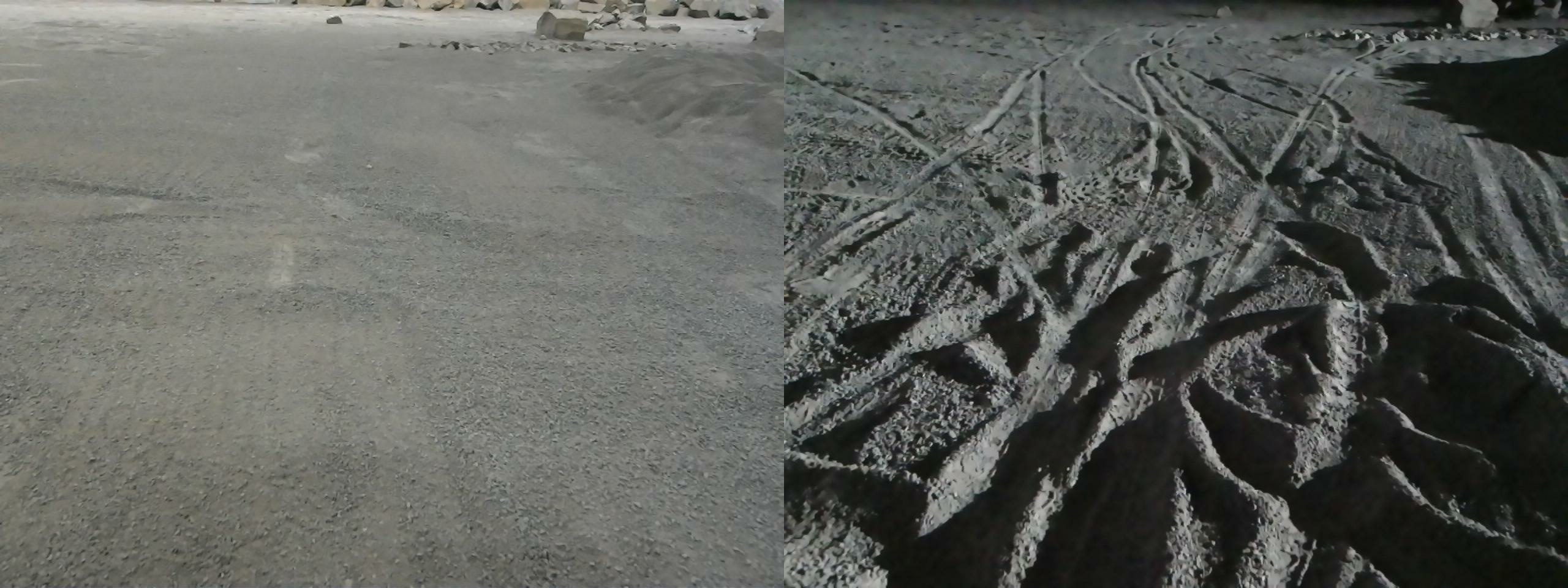}
    
    \vspace{2mm}
    
    \includegraphics[width=\columnwidth]{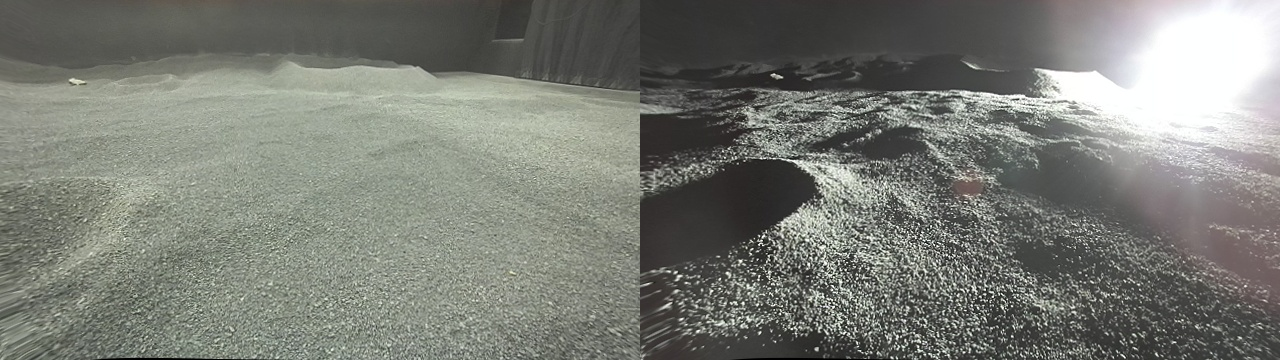}
    
    \caption{Examples from the two analogue datasets under different lighting conditions. Top: In-house dataset; bottom: LunaPolaris dataset..}
    \label{fig:dataset_lighting_examples}

\end{figure}

For cross-dataset lunar traversability estimation, three datasets are used, covering two lunar analogue test facilities and real lunar imagery from rover missions:

\begin{itemize}
    \item \emph{LunaPolaris}~\cite{van2026lunapolaris} was collected in the LunaLab~\cite{ludivig2020building} with a small rover. This indoor lunar analogue facility provides controlled regolith-like terrain under repeatable lighting conditions. RGB images were collected using a ZED2 stereo camera. Because the authors provide synchronised left and right images for each timestamp, along with calibration data, we use standard stereo vision techniques to compute depth maps. Figure~\ref{fig:dataset_lighting_examples} shows that the data used were taken under both diffuse and harsh lighting conditions.

    \item \emph{In-House dataset} refers to our own analogue test facility at OTRK Kakolewo\footnote{\url{http://otrk.put.poznan.pl/}}, equipped with diverse rock formations and crater-like features designed to emulate visual characteristics of lunar surface morphology. We collected visual data and depth images using an Orbbec Femto Mega camera mounted on a mobile robot. The camera utilises an indirect Time-of-Flight approach, emitting continuously modulated light and computing depth from the phase shift between emitted and reflected signals, yielding dense depth maps; however, its performance degrades under strong ambient sunlight. We collected data across varied lighting conditions, including low light, full daylight, and low solar angles that produce strong shadows and lens flares. Figure~\ref{fig:dataset_lighting_examples} also shows that the data were gathered in both diffuse and harsh lighting conditions, similar to LunaPolaris.

    \item \emph{CE’3/CE’4}. We use real lunar surface imagery captured by the panoramic cameras (PCAM) of the Chang’E-3~\cite{gras_ce3_pcam} and Chang’E-4~\cite{gras_ce4_pcam} missions, obtained via the Ground Research and Application System (GRAS) of the China Lunar Exploration Program~\cite{zuo2021clpds}. The PCAM instruments aboard the Yutu and Yutu-2 rovers acquired ground-level colour images of the lunar surface in Mare Imbrium and the Von Kármán crater on the lunar farside, respectively. Although stereo image pairs are available for the panoramic cameras, they are not temporally aligned. Moreover, calibration data from the rover missions are not publicly available. Consequently, we had to make several assumptions for stereo reconstruction. First, we used the camera parameters reported in the datasheet~\cite{yang2015panoramic} as an initial estimate. Second, to rectify unsynchronised image pairs, we applied a point-based alignment approach. Specifically, we detected keypoints using SuperPoint~\cite{detone2018superpoint} and matched them with LightGlue~\cite{lindenberger2023lightglue} to estimate a transformation between image pairs, without considering real-time constraints. Third, to address appearance differences, we used the S2M2 method~\cite{min2025s2m2}, a global matching neural network, to compute disparity. Repeated size estimation experiments on the Chinese national flag visible on the lander yielded a consistent error of approximately 15\%, which is acceptable given the absence of precise calibration.

\end{itemize}

\noindent
For each dataset, manual annotations were prepared based on sparse pairwise comparisons following the W-RIZZ protocol~\cite{schreiber2024w}. For each image, an annotator labels one of the point pairs as \emph{more traversable}, \emph{less traversable}, or \emph{equally traversable}. The process does not define levels of traversability; rather, it ranks them, defining a global relation between image pixels. We select labelling points arbitrarily, focusing on certain relations and uncommon cases. In addition, inter-image pairwise comparisons are included, which improve the consistency of predictions across different scenes~\cite{schreiber2025you}. A quantitative comparison of the datasets, along with their characteristics, is provided in Table~\ref{table:dataset_comparison}.

\subsection{Metrics}

We adopt the Human Disagreement Rate (HDR) metric introduced in W-RIZZ~\cite{schreiber2024w}, which measures the fraction of annotated point pairs for which the model's predicted ordinal relation disagrees with the human label, using a threshold $t$ on the predicted score difference. Unlike absolute measures like MSE, HDR evaluates the relative ranking of traversability between points, aligning with the weakly supervised pairwise annotation scheme and accommodating the intrinsic uncertainty in assigning traversability scores. We report HDR averaged over thresholds $t \in [0.10, 0.50]$ as the main metric, along with the standard deviation, with HDR at $t=0.25$ for direct comparison with prior work. To reduce the influence of randomness, we ran all experiments 10 times with different seeds and report the mean and standard deviation.

\subsection{Results}

\subsubsection{Quantitative}

\begin{table}[htbp]
    \centering
    \caption{Results of traversability evaluation among datasets, averaged over 10 random seeds.}
    \label{table:metric_results}
    \setlength{\tabcolsep}{2pt}
    \scriptsize
    \vspace{0.2cm}

    \begin{tabular}{lllcc}
        \toprule
        \textbf{Train Data}
        & \textbf{ Test Data}
        & \textbf{Mode}
        & \shortstack{\textbf{HDR@0.25 $\downarrow$}}
        & \shortstack{\textbf{Avg. HDR $\downarrow$}} \\
        \midrule

        \multirow{4}{*}{In-house}
        & \multirow{4}{*}{In-house}
        & \cellcolor{visualcolor}Visual
        & \cellcolor{visualcolor}0.030 $\pm$ 0.003
        & \cellcolor{visualcolor}0.043 $\pm$ 0.003 \\

        & &
        \cellcolor{depthcolor}Depth
        & \cellcolor{depthcolor}0.030 $\pm$ 0.002
        & \cellcolor{depthcolor}0.042 $\pm$ 0.002 \\

        & &
        \cellcolor{normalscolor}Normals
        & \cellcolor{normalscolor}0.031 $\pm$ 0.003
        & \cellcolor{normalscolor}0.042 $\pm$ 0.002 \\

        & &
        \cellcolor{geometriccolor}Geometric
        & \cellcolor{geometriccolor}0.023 $\pm$ 0.002
        & \cellcolor{geometriccolor}0.043 $\pm$ 0.002 \\
        \addlinespace

        \multirow{4}{*}{LunaPolaris}
        & \multirow{4}{*}{LunaPolaris}
        & \cellcolor{visualcolor}Visual
        & \cellcolor{visualcolor}0.046 $\pm$ 0.007
        & \cellcolor{visualcolor}0.056 $\pm$ 0.007 \\

        & &
        \cellcolor{depthcolor}Depth
        & \cellcolor{depthcolor}0.041 $\pm$ 0.007
        & \cellcolor{depthcolor}0.053 $\pm$ 0.007 \\

        & &
        \cellcolor{normalscolor}Normals
        & \cellcolor{normalscolor}0.040 $\pm$ 0.005
        & \cellcolor{normalscolor}0.051 $\pm$ 0.005 \\

        & &
        \cellcolor{geometriccolor}Geometric
        & \cellcolor{geometriccolor}0.040 $\pm$ 0.003
        & \cellcolor{geometriccolor}0.052 $\pm$ 0.005 \\
        \addlinespace

        \multirow{4}{*}{In-house}
        & \multirow{4}{*}{LunaPolaris}
        & \cellcolor{visualcolor}Visual
        & \cellcolor{visualcolor}0.119 $\pm$ 0.013
        & \cellcolor{visualcolor}0.145 $\pm$ 0.010 \\

        & &
        \cellcolor{depthcolor}Depth
        & \cellcolor{depthcolor}0.107 $\pm$ 0.016
        & \cellcolor{depthcolor}0.135 $\pm$ 0.015 \\

        & &
        \cellcolor{normalscolor}Normals
        & \cellcolor{normalscolor}0.115 $\pm$ 0.009
        & \cellcolor{normalscolor}0.138 $\pm$ 0.010 \\

        & &
        \cellcolor{geometriccolor}Geometric
        & \cellcolor{geometriccolor}0.114 $\pm$ 0.007
        & \cellcolor{geometriccolor}0.139 $\pm$ 0.007 \\
        \addlinespace

        \multirow{4}{*}{LunaPolaris}
        & \multirow{4}{*}{In-house}
        & \cellcolor{visualcolor}Visual
        & \cellcolor{visualcolor}0.075 $\pm$ 0.014
        & \cellcolor{visualcolor}0.091 $\pm$ 0.012 \\

        & &
        \cellcolor{depthcolor}Depth
        & \cellcolor{depthcolor}0.077 $\pm$ 0.007
        & \cellcolor{depthcolor}0.093 $\pm$ 0.005 \\

        & &
        \cellcolor{normalscolor}Normals
        & \cellcolor{normalscolor}0.079 $\pm$ 0.011
        & \cellcolor{normalscolor}0.095 $\pm$ 0.011 \\

        & &
        \cellcolor{geometriccolor}Geometric
        & \cellcolor{geometriccolor}0.072 $\pm$ 0.007
        & \cellcolor{geometriccolor}0.089 $\pm$ 0.007 \\
        \addlinespace

        \multirow{4}{*}{In-house}
        & \multirow{4}{*}{CE'3/CE'4}
        & \cellcolor{visualcolor}Visual
        & \cellcolor{visualcolor}0.153 $\pm$ 0.006
        & \cellcolor{visualcolor}0.167 $\pm$ 0.008 \\

        & &
        \cellcolor{depthcolor}Depth
        & \cellcolor{depthcolor}0.150 $\pm$ 0.015
        & \cellcolor{depthcolor}0.168 $\pm$ 0.015 \\

        & &
        \cellcolor{normalscolor}Normals
        & \cellcolor{normalscolor}0.153 $\pm$ 0.008
        & \cellcolor{normalscolor}0.166 $\pm$ 0.007 \\

        & &
        \cellcolor{geometriccolor}Geometric
        & \cellcolor{geometriccolor}0.154 $\pm$ 0.011
        & \cellcolor{geometriccolor}0.168 $\pm$ 0.009 \\
        \addlinespace

        \multirow{4}{*}{LunaPolaris}
        & \multirow{4}{*}{CE'3/CE'4}
        & \cellcolor{visualcolor}Visual
        & \cellcolor{visualcolor}0.102 $\pm$ 0.012
        & \cellcolor{visualcolor}0.118 $\pm$ 0.011 \\

        & &
        \cellcolor{depthcolor}Depth
        & \cellcolor{depthcolor}0.105 $\pm$ 0.014
        & \cellcolor{depthcolor}0.123 $\pm$ 0.011 \\

        & &
        \cellcolor{normalscolor}Normals
        & \cellcolor{normalscolor}0.103 $\pm$ 0.017
        & \cellcolor{normalscolor}0.120 $\pm$ 0.016 \\

        & &
        \cellcolor{geometriccolor}Geometric
        & \cellcolor{geometriccolor}0.098 $\pm$ 0.010
        & \cellcolor{geometriccolor}0.118 $\pm$ 0.009 \\
        \addlinespace

        \multirow{4}{*}{\shortstack{LunaPolaris\\+ In-house}}
        & \multirow{4}{*}{CE'3/CE'4}
        & \cellcolor{visualcolor}Visual
        & \cellcolor{visualcolor}0.107 $\pm$ 0.008
        & \cellcolor{visualcolor}0.125 $\pm$ 0.006 \\

        & &
        \cellcolor{depthcolor}Depth
        & \cellcolor{depthcolor}0.113 $\pm$ 0.009
        & \cellcolor{depthcolor}0.128 $\pm$ 0.008 \\

        & &
        \cellcolor{normalscolor}Normals
        & \cellcolor{normalscolor}0.109 $\pm$ 0.012
        & \cellcolor{normalscolor}0.129 $\pm$ 0.008 \\

        & &
        \cellcolor{geometriccolor}Geometric
        & \cellcolor{geometriccolor}0.124 $\pm$ 0.012
        & \cellcolor{geometriccolor}0.138 $\pm$ 0.009 \\

        \bottomrule
    \end{tabular}
\end{table}

\begin{figure}[ht!]
  \centering
  \includegraphics[width=\linewidth]{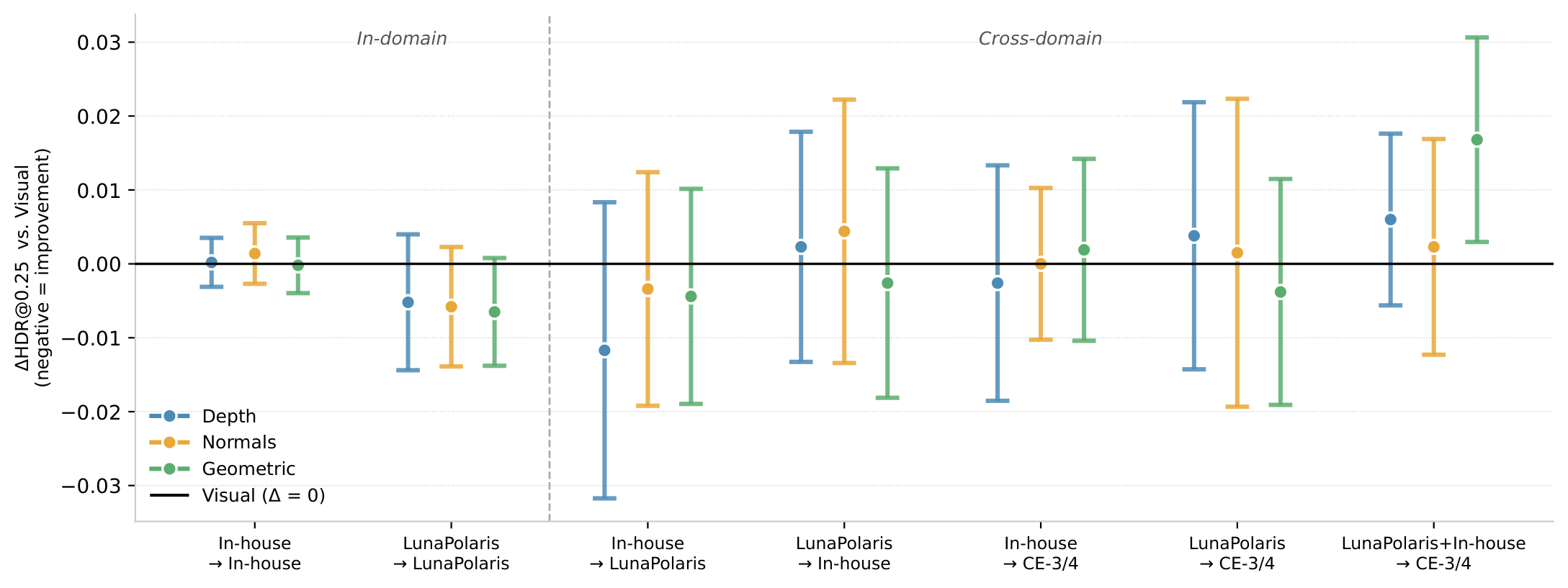}
  \caption{Change in HDR@0.25 relative to the visual-only baseline for each geometric fusion mode (Depth, Normals, Geometric). Values below zero indicate improvement over the visual baseline; values above zero indicate degradation.}
  \label{fig:fusion_delta}
\end{figure}

Table~\ref{table:metric_results} reports full numeric results of HDR@0.25 and average HDR across all train--test domain pairs and input modalities, while Figure~\ref{fig:fusion_delta} summarises the change in HDR@0.25 relative to the visual-only baseline across all train/test combinations and geometric fusion modes.

In in-domain settings, geometric fusion has a small effect throughout. For \emph{$\text{In-house} \rightarrow \text{In-house}$}, all three modes remain within one standard deviation of the visual baseline, with no mode showing a statistically distinguishable improvement. For \emph{$\text{LunaPolaris} \rightarrow \text{LunaPolaris}$}, Depth and Normals yield modest gains ($\Delta$ HDR@0.25 of -0.005 and -0.006, respectively), while Geometric produces the largest improvement of the two in-domain cases (-0.006, HDR@0.25 0.040 vs 0.046 for Visual).

In cross-domain transfer, the results are inconsistent across both settings and modalities. The clearest benefit appears for \emph{$\text{In-house} \rightarrow \text{LunaPolaris}$}, where Depth reduces HDR@0.25 from 0.119 to 0.107 ($\Delta$ = -0.012), the largest absolute improvement observed. For \emph{$\text{LunaPolaris} \rightarrow \text{In-house}$}, Geometric achieves the best result (0.072 vs. 0.075 for Visual, $\Delta$ = -0.003), while Depth and Normals slightly degrade performance. Transfer to CE-3/4 shows no consistent pattern: for \emph{$\text{In-house} \rightarrow \text{CE-3/4}$}, all modes perform within the noise of Visual, and for \emph{$\text{LunaPolaris}+\text{In-house} \rightarrow \text{CE-3/4}$}, Geometric is the worst-performing mode (0.124 vs 0.107 for Visual, $\Delta$ = +0.017), the only case where geometric fusion causes a clearly detrimental effect.

Overall, no modality consistently outperforms the visual-only baseline across the evaluated settings. Where geometric fusion helps, the improvements are small relative to the seed-to-seed variance, and in the one case where it clearly changes the result -- \emph{$\text{LunaPolaris}+\text{In-house} \rightarrow \text{CE-3/4}$} -- the effect is detrimental. We therefore do not claim that geometric fusion improves traversability estimation under the naive concatenation strategy used here; rather, these results establish a controlled baseline against which more expressive multimodal architectures can be compared in future work.

We note that the apparent benefit of geometric augmentation is substantially smaller when averaged over multiple seeds than under a single training run: effects that appear sizeable in an individual run (exceeding 30\% relative HDR reduction in some transfers) fall within seed-to-seed variance once averaged over 10 runs. This reinforces the importance of multi-seed evaluation when comparing closely-performing modalities.

\subsubsection{Visual}

\begin{figure*}[p] 
    \centering
    
    \begin{subfigure}{0.4\textwidth}
        \includegraphics[width=\linewidth]{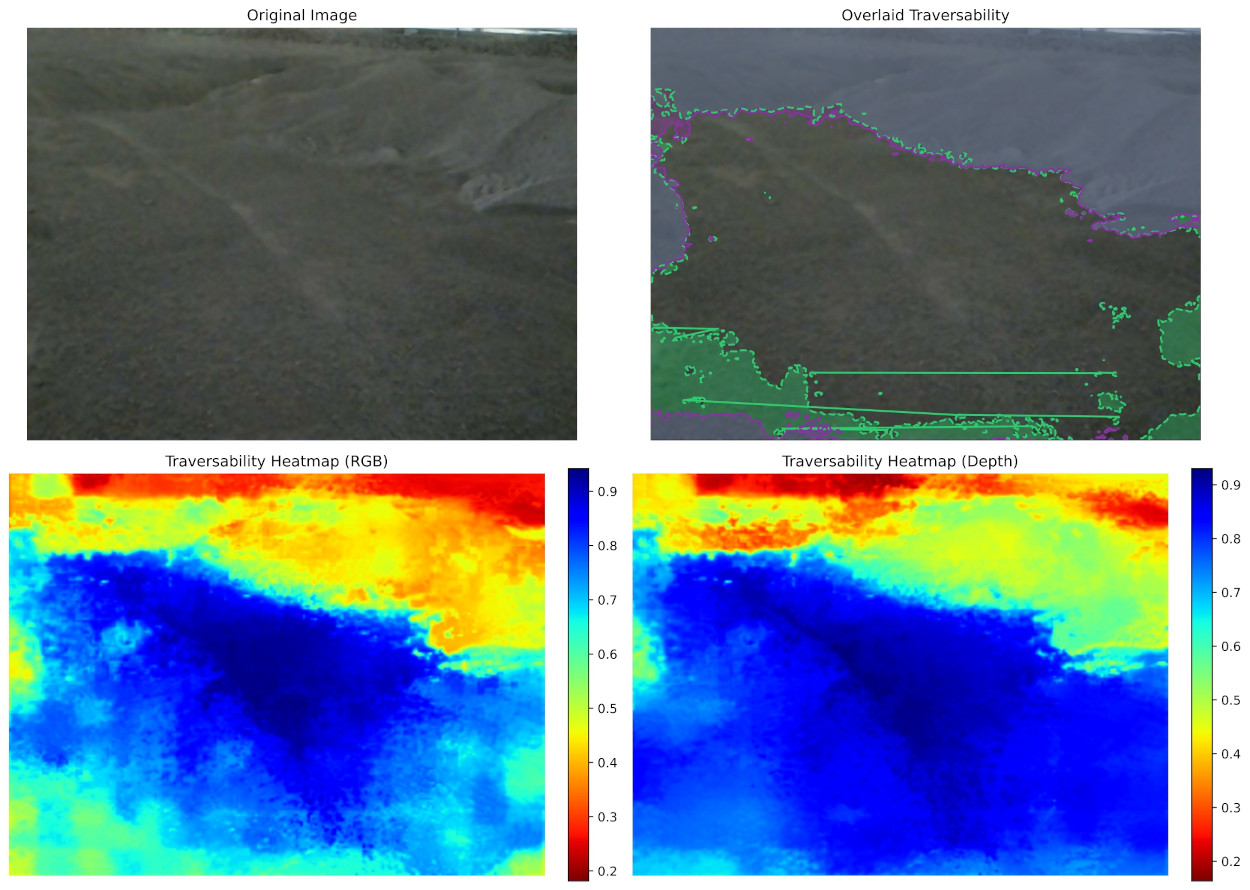}
        \caption{In-domain: In-House $\rightarrow$ In-House}
        \label{fig:img1}
    \end{subfigure}\hfill
    \begin{subfigure}{0.4\textwidth}
        \includegraphics[width=\linewidth]{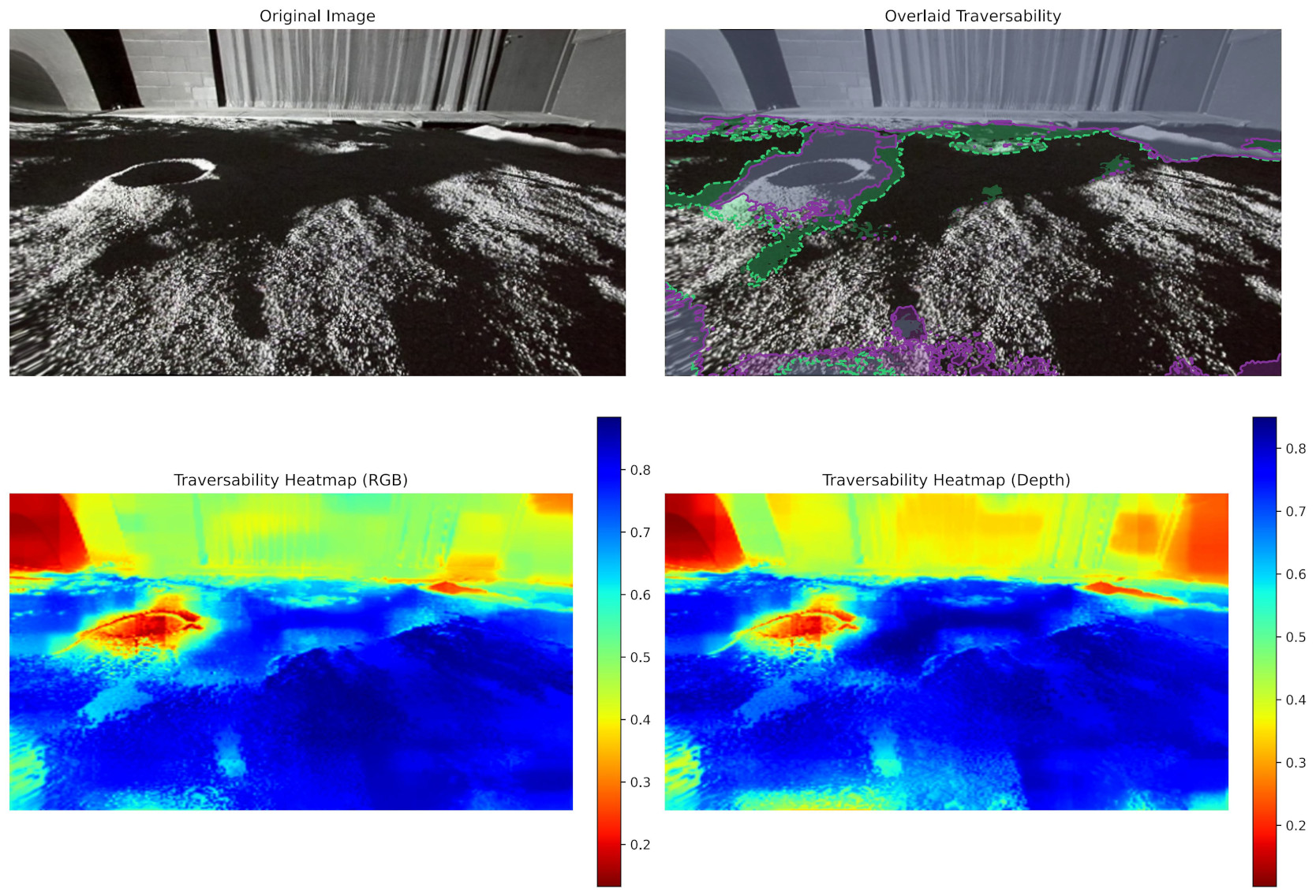}
        \caption{In-domain: LunaPolaris $\rightarrow$ LunaPolaris}
        \label{fig:img3}
    \end{subfigure}

    \vspace{0.1cm} 
    
    \begin{subfigure}{0.4\textwidth}
        \includegraphics[width=\linewidth]{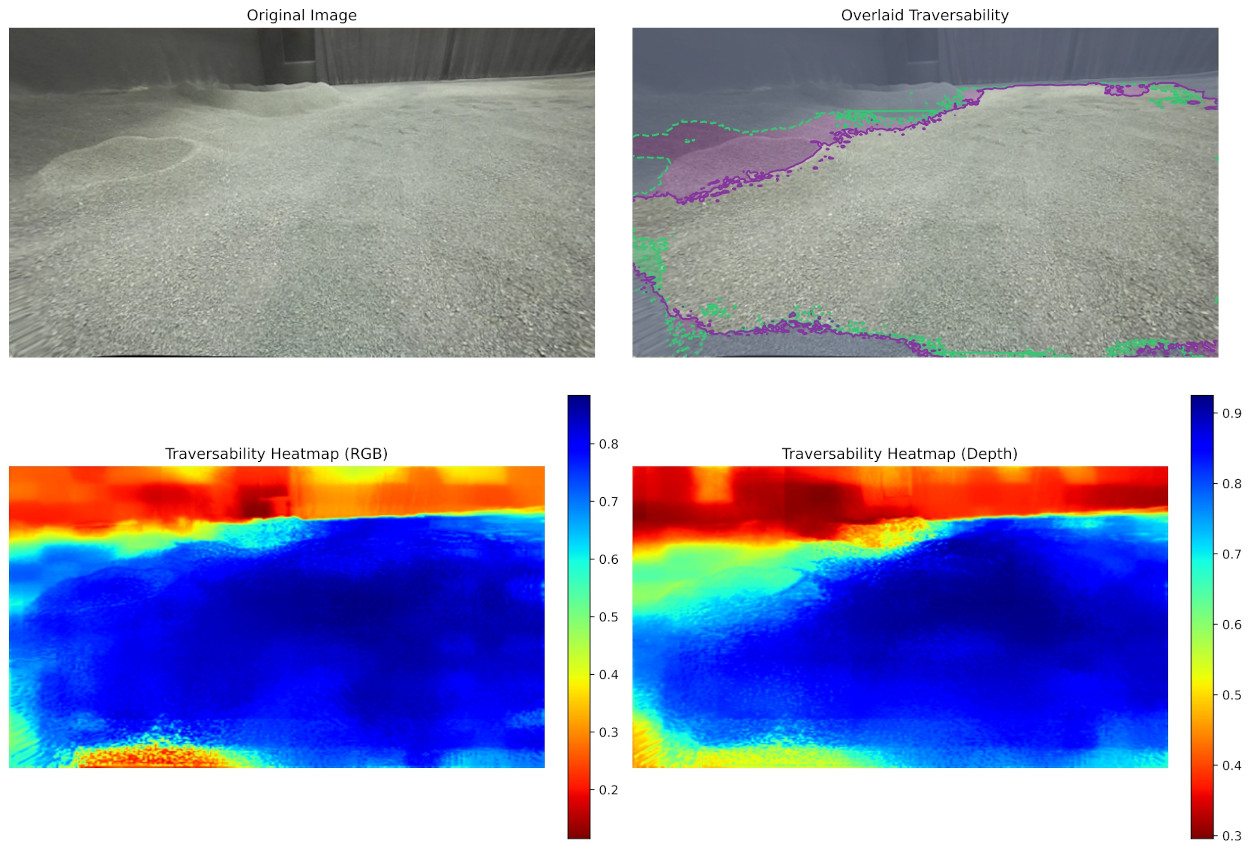}
        \caption{Cross-domain: In-House $\rightarrow$ LunaPolaris}
        \label{fig:img4}
        \end{subfigure}\hfill
    \begin{subfigure}{0.4\textwidth}
        \includegraphics[width=\linewidth]{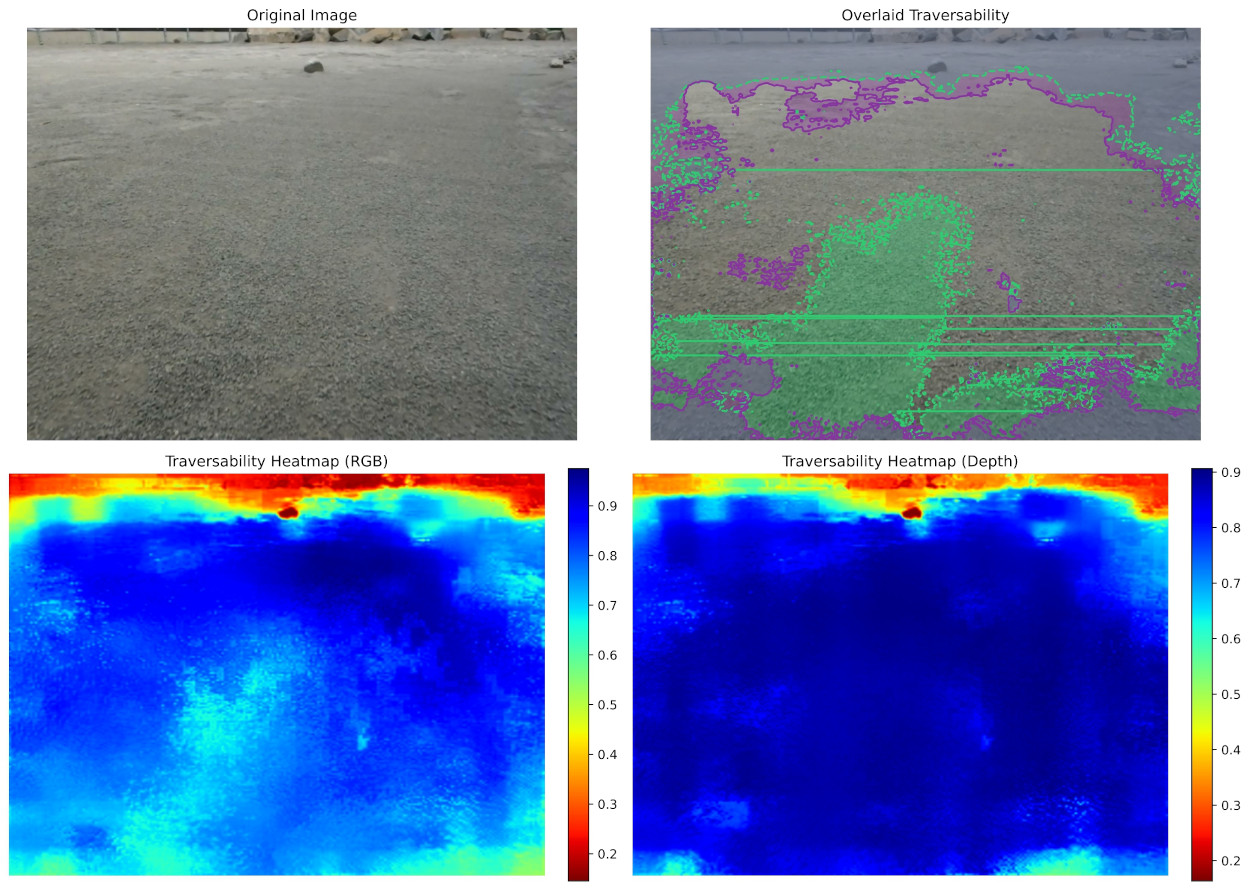}
        \caption{Cross-domain: LunaPolaris $\rightarrow$ In-House}
        \label{fig:img2}
    \end{subfigure}
    
    \vspace{0.1cm}
    
    \begin{subfigure}{0.4\textwidth}
        \includegraphics[width=\linewidth]{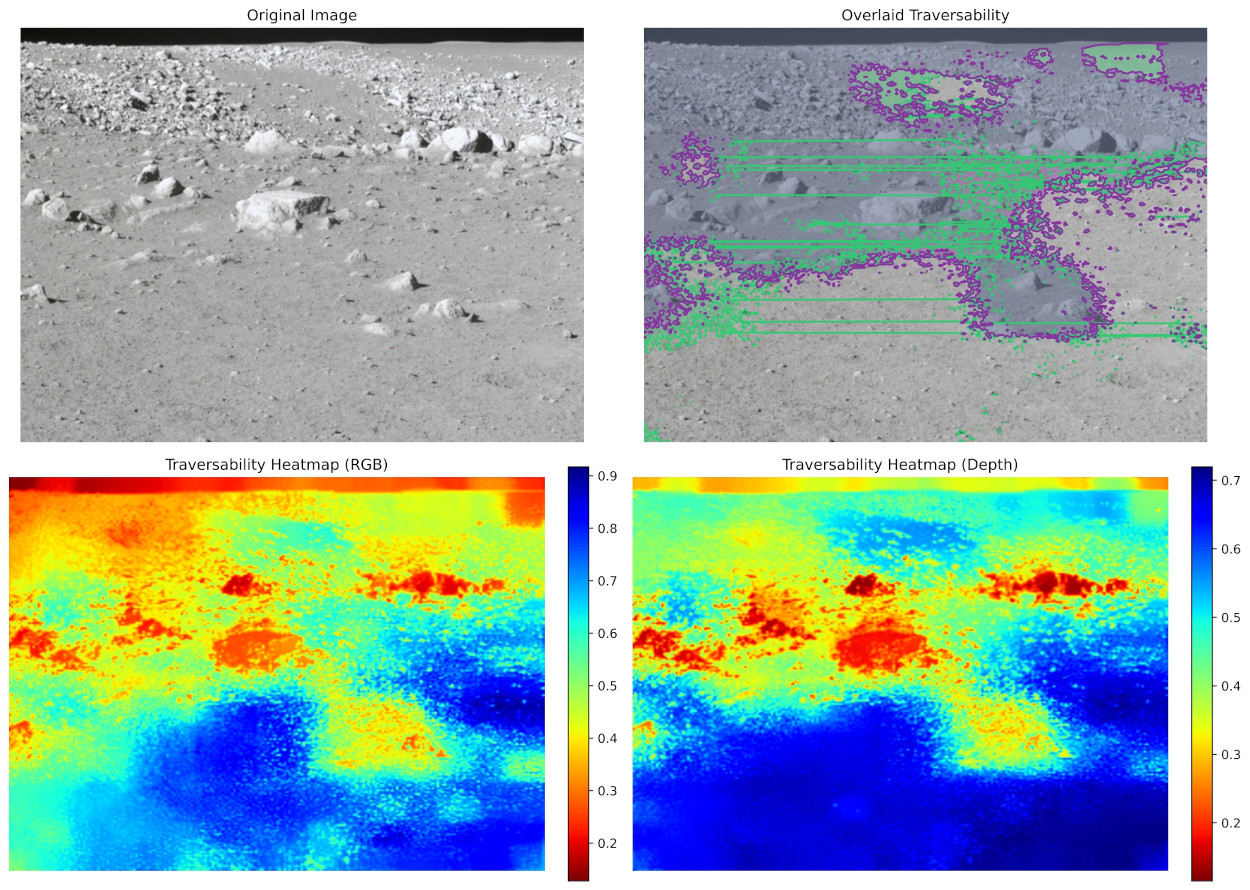}
        \caption{Cross-domain: In-House $\rightarrow$ Chang'E-3/4}
        \label{fig:img5}
    \end{subfigure}\hfill
    \begin{subfigure}{0.4\textwidth}
        \includegraphics[width=\linewidth]{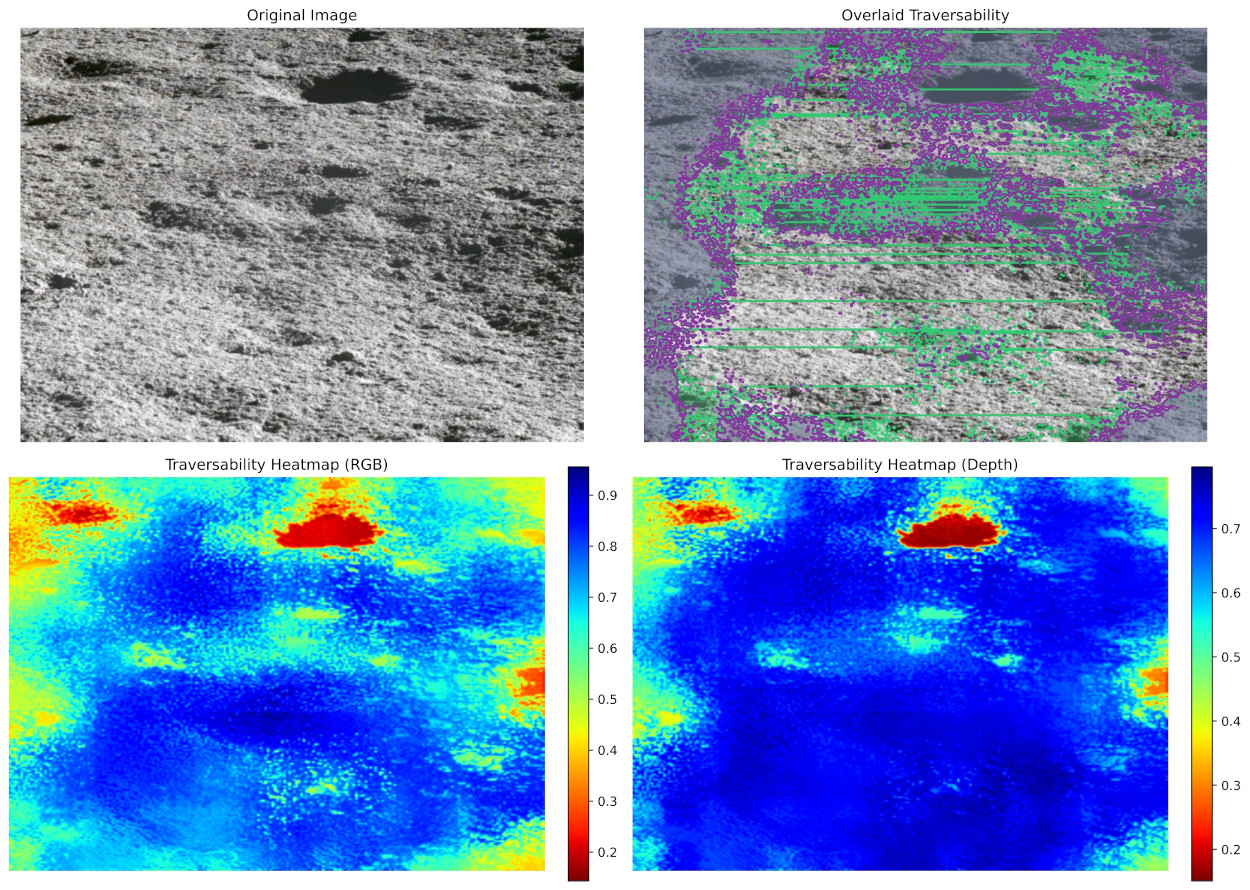}
        \caption{Cross-domain: LunaPolaris $\rightarrow$ Chang'E-3/4}
        \label{fig:img6}
    \end{subfigure}
    
    \vspace{0.1cm}
    
    \begin{subfigure}{0.8\textwidth} 
        \centering 
        \includegraphics[width=0.5\linewidth]{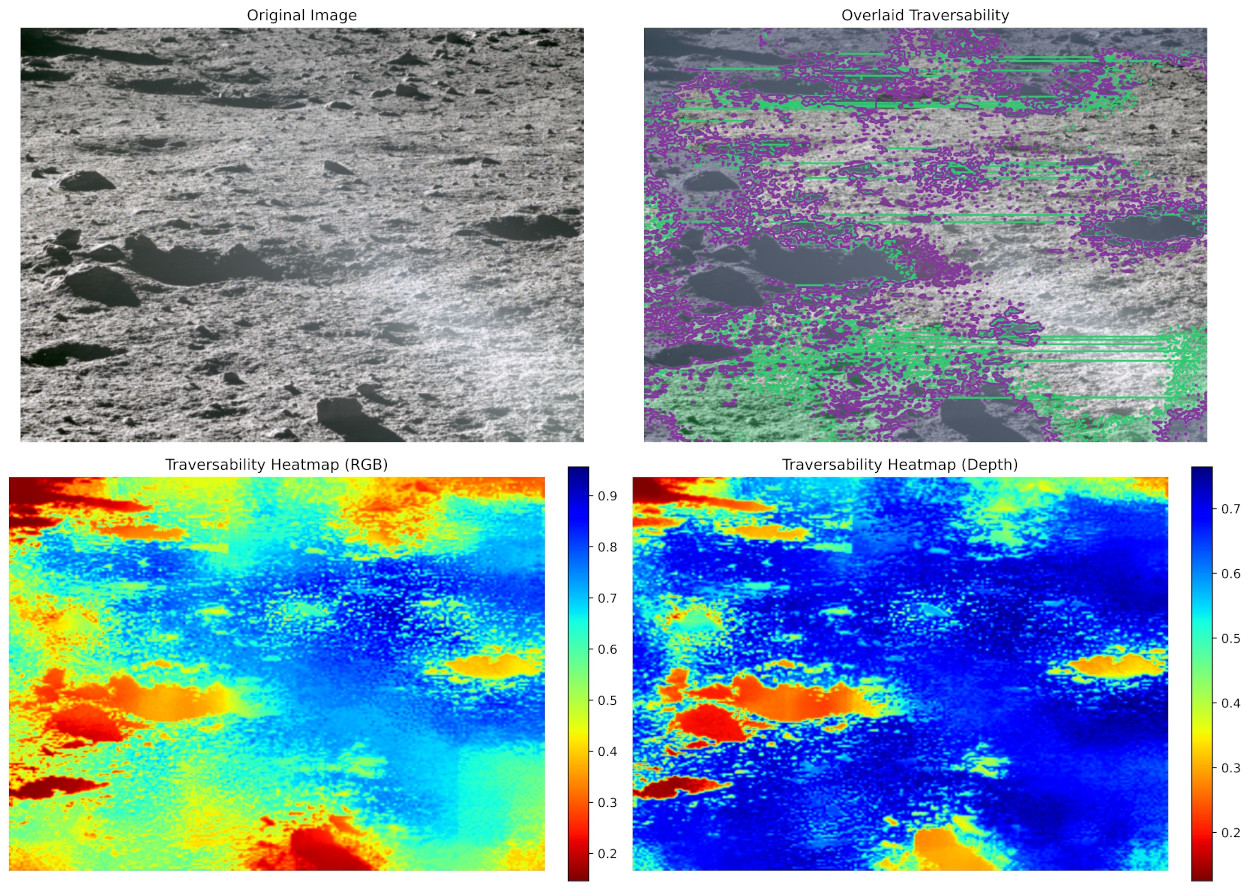} 
        \caption{Combined source: In-House + LunaPolaris $\rightarrow$ Chang'E-3/4}
        \label{fig:img7}
    \end{subfigure}
    
    \caption{Qualitative traversability predictions across all train-test domain pairs. Traversability heatmaps: warmer colours indicate lower scores. Overall traversability: green - visual is more sensitive, purple - Depth is more sensitive.}
    \label{fig:all_scenarios}
\end{figure*}

Figure~\ref{fig:all_scenarios} provides qualitative traversability predictions for all train--test pairs. To further highlight the differences between prediction outputs, we provide an additional qualitative visualisation. Since the traversability score distributions differ across methods, we first discard the top and bottom 5\% of prediction values for each method. We then compute the mean of the remaining values and use it as a threshold to separate traversable and non-traversable regions. We overlay the resulting contours on the original RGB image, with coloured regions indicating areas classified as non-traversable. Visual-only-based predictions are shown in green, while depth-based predictions are shown in purple. In in-domain settings (Figs.~\ref{fig:all_scenarios}a--b), both Visual and Depth heatmaps produce spatially coherent predictions that align well with annotated boundaries: flat regolith receives high traversability scores while rocks and crater features are correctly penalised. The differences are particularly visible in cross-domain evaluations on terrestrial analogue datasets. In some cases, the Visual model fails to capture subtle variations between terrain types, leading to uncertain or inconsistent predictions. For example, it may classify craters as highly traversable or flat surfaces as poorly traversable (Figs.~\ref{fig:all_scenarios}c--d). For cross-domain evaluation on real lunar data, the visual-based predictions are generally spatially plausible, but the depth-enhanced models produce more cohesive regions and show greater confidence in distinguishing traversable from non-traversable terrain (Figs.~\ref{fig:all_scenarios}e--g).

\subsubsection{Edge Deployment Evaluation}

To assess edge deployment feasibility, we evaluated each model variant on an NVIDIA Jetson AGX Orin (JetPack 6.2.2, L4T 36.5.0) using ONNX Runtime 1.22.0 with four execution configurations: CPU, CUDA, TensorRT FP32, and TensorRT FP16. Measured execution time included both model inference and modality-specific preprocessing (depth inpainting and geometric descriptor computation). Throughput was reported as mean ± standard deviation over 100 inferences, expressed in frames per second (FPS).

Table~\ref{tab:inference_throughput} presents the throughput obtained for four model variants and four execution modes. CPU-only execution achieved approximately $0.21$--$0.22$~FPS and was unsuitable for real-time operation. CUDA acceleration increased throughput to $13.7$--$17.1$~FPS, while TensorRT produced between $18.7$ and $29.1$~FPS. The visual and depth variants reached almost $29$~FPS with TensorRT, demonstrating the feasibility of near-real-time edge deployment.

The differences between the TensorRT FP32 and FP16 configurations were small relative to the measured variability. FP16 achieved slightly higher mean throughput for the visual, depth, and normals variants, whereas FP32 was marginally faster for the Geometric variant. Therefore, reduced numerical precision did not consistently improve throughput, and you should also consider its effect on prediction quality when selecting the deployment configuration.

The visual and depth variants obtained similar throughput because they require only basic preprocessing. Normals and Geometric achieved lower throughput because they must compute surface normals and additional terrain descriptors before inference. The results therefore show a trade-off between the information provided by geometric representations and the operating frequency of the full perception pipeline.

\begin{table}[t]
    \centering
    \caption{Inference throughput for different models and execution modes.}
    \label{tab:inference_throughput}
    \begin{tabular}{llc}
        \toprule
        \textbf{Mode} & \textbf{Execution mode}
        & \textbf{Throughput [FPS]} $\uparrow$ \\
        \midrule

        \multirow{4}{*}{Visual}
        & CPU            & $0.215 \pm 0.001$ \\
        & CUDA           & $17.09 \pm 1.02$ \\
        & TensorRT FP32  & $28.49 \pm 2.53$ \\
        & TensorRT FP16  & $28.93 \pm 2.32$ \\
        \midrule

        \multirow{4}{*}{Depth}
        & CPU            & $0.219 \pm 0.001$ \\
        & CUDA           & $16.74 \pm 1.19$ \\
        & TensorRT FP32  & $27.34 \pm 2.33$ \\
        & TensorRT FP16  & $29.14 \pm 2.45$ \\
        \midrule

        \multirow{4}{*}{Normals}
        & CPU            & $0.211 \pm 0.001$ \\
        & CUDA           & $14.09 \pm 0.65$ \\
        & TensorRT FP32  & $19.35 \pm 1.21$ \\
        & TensorRT FP16  & $20.23 \pm 2.17$ \\
        \midrule

        \multirow{4}{*}{Geometric}
        & CPU            & $0.214 \pm 0.001$ \\
        & CUDA           & $13.73 \pm 0.58$ \\
        & TensorRT FP32  & $19.10 \pm 1.38$ \\
        & TensorRT FP16  & $18.65 \pm 1.99$ \\
        \bottomrule
    \end{tabular}
\end{table}


\section{Conclusions}
We presented a cross-dataset evaluation of geometry-augmented DINOv2 features for lunar traversability estimation. Our method extends the weakly supervised CHUNGUS framework by concatenating depth-derived geometric channels -- depth, surface normals, and handcrafted terrain descriptors -- with AnyUp-upsampled DINOv2 embeddings before the traversability prediction head. Experiments across two analogue facilities and real Chang'E-3/4 lunar imagery show that, under this naive concatenation strategy, geometric augmentation has a negligible and inconsistent effect in both in-domain and cross-domain settings: where improvements occur, they are small relative to seed-to-seed variance, and in one cross-domain case geometric fusion is clearly detrimental. We therefore treat these results not as evidence that geometry improves traversability estimation, but as a controlled baseline that isolates the marginal contribution of geometric cues over a strong visual-only model and against which future work can compare more expressive multimodal fusion architectures. Independently of fusion mode, the choice of training source has a larger effect on transfer quality: LunaPolaris proves a more effective training base for real-lunar deployment than our In-House facility, consistent with the broader finding that domain relevance, rather than sensing modality, is the dominant factor in cross-domain traversability transfer.

\subsection{Limitations}

The study has several limitations; we highlight two key issues. First, the handcrafted pairwise annotations were produced by a single annotator, which may introduce subjective bias and limit the assessment of traversability to the selected image regions and their visual characteristics. Second, the depth data across the evaluated datasets were obtained using different sensing modalities, including conventional stereo vision, deep learning-enhanced stereo vision, and indirect time-of-flight. Consequently, the observed generalisation of depth-based representations may partially reflect not only geometric invariance but also appearance normalisation introduced by the respective depth estimation pipelines.

\subsection{Future work}
Future work will extend the evaluation by examining model robustness under varying lighting conditions and systematically comparing alternative methods of incorporating geometric information into the model. We also plan to shift the focus of this research from algorithm-centric evaluation toward robot-centric validation. This includes integrating the proposed method with a path-planning algorithm and deploying it on a real rover in our in-house OTRK facility, where performance will be assessed using metrics such as trajectory length, obstacle avoidance, and power consumption. Another direction for future research is to investigate how visual traversability information can support teleoperation by analysing its impact on operator behaviour, navigation accuracy, and overall task performance.

\section*{DATA ACCESS}

The dataset (colour images, depth maps, camera parameters, and annotations) collected in our lunar facility is publicly available here: \url{https://doi.org/10.5281/zenodo.22871299}.

\section*{DISCLOSURE}

The authors acknowledge the use of a large language model solely for editing and enhancing the manuscript's grammar. No sections of the article, including the main technical content, figures, or images, were generated by the AI system. The use was limited to improving readability and clarity of language.

\section*{ACKNOWLEDGMENT}

Jan Baturo's work was supported by the European Space Agency (ESA Spaceship Poland).

\bibliographystyle{IEEEtran}
\bibliography{IEEEexample}

\end{document}